\documentclass[letterpaper, 10 pt, conference]{ieeeconf}  

\IEEEoverridecommandlockouts                              

\usepackage{graphicx}
\usepackage{caption}
\usepackage{subcaption}
\usepackage{etoolbox}
\usepackage{lipsum}
\usepackage{amssymb}
\makeatletter
\patchcmd{\@verbatim}
  {\verbatim@font}
  {\verbatim@font\small}
  {}{}
\makeatother

\usepackage{algorithm}
\usepackage[noend]{algpseudocode}
\usepackage{blindtext}
\algnewcommand\INPUT{\item[\textbf{Input:}]}%
\algnewcommand\OUTPUT{\item[\textbf{Output:}]}%
\newcommand{\algorithmicbreak}{\textbf{break}}
\newcommand{\Break}{\State \algorithmicbreak}
\newcommand{\algorithmicpass}{\textbf{continue}}
\newcommand{\CONTINUE}{\algorithmicpass}
\usepackage{amsmath}
\DeclareMathOperator*{\argmin}{\mathbf{argmin}}
\long\def\/*#1*/{}

\title{\LARGE \bf
Graph-based Topological Exploration Planning in Large-scale 3D Environments  
}

\author{Fan Yang, Dung-Han Lee, John Keller and Sebastian Scherer 
\thanks{*All authors are with the Robotics Institue at Carnegie Mellon University, Pittsburgh. Emails: \{ {\tt\small fanyang2, dunghanl, jkeller2, basti} \} {@andrew.cmu.edu} }
}

\begin{document}

\maketitle
\thispagestyle{empty}
\pagestyle{empty}


\begin{abstract}
Currently, state-of-the-art exploration methods maintain high-resolution map representations in order to optimize exploration goals in each step that maximizes information gain. However, during exploring, those ``optimal" selections could quickly become obsolete due to the influx of new information, especially in large-scale environments, and result in high-frequency re-planning that hinders the overall exploration efficiency. In this paper, we propose a graph-based topological planning framework, building a sparse topological map in three-dimensional (3D) space to guide exploration steps with high-level intents so as to render consistent exploration maneuvers. Specifically, this work presents a novel method to estimate 3D space's geometry with convex polyhedrons. Then, the geometry information is utilized to group space into distinctive regions. And those regions are added as nodes into the topological map, directing the exploration process. We compared our method with the state-of-the-art in simulated environments. The proposed method achieves higher space coverage and outperforms exploration efficiency by more than $\mathbf{40\%}$ during experiments. Finally, a field experiment was conducted to further evaluate the applicability of our method to empower efficient and robust exploration in real-world environments.
\end{abstract}


\section{Introduction}

Exploration is a classic problem in the field of robotics. It is relevant to applications that are too hazardous or costly for humans to operate in e.g. disaster response \cite{Thakur2018NuclearEI}, search and rescue operations \cite{6696431}, \cite{20.500.11850/116421}. One particular case that has gained attention from recent research is underground exploration, \cite{8981594}, \cite{8968151}, \cite{8797885} which arose from the DARPA Subterranean Challenge. These underground scenarios pose various challenges to robot exploration such as GPS-denial, dark and textureless environments with large, multi-branched space that stretch for kilometers. In those scenarios, aerial vehicles gain favours over ground vehicles because of the presence of challenging terrains \cite{8981594}, \cite{6225146}. However, due to the constrained battery lifetime and the sheer scale of the environments, it’s even more pressing for the aerial vehicles to improve the exploration efficiency and lower the computational cost. To this end, we propose a method to estimate and separate regions in 3D space with convex polyhedrons, and build a topological representation of the 3D environment based on their geometry contiguity and mutual visibility. Then, a local and global planning strategy are employed in a complementary fashion on the topological map to direct the exploration process. The proposed method is compared in simulated environments with method \cite{8968151}, considered as state-of-the-art. 

\begin{figure}[h!]
\centering
\includegraphics[scale=0.33]{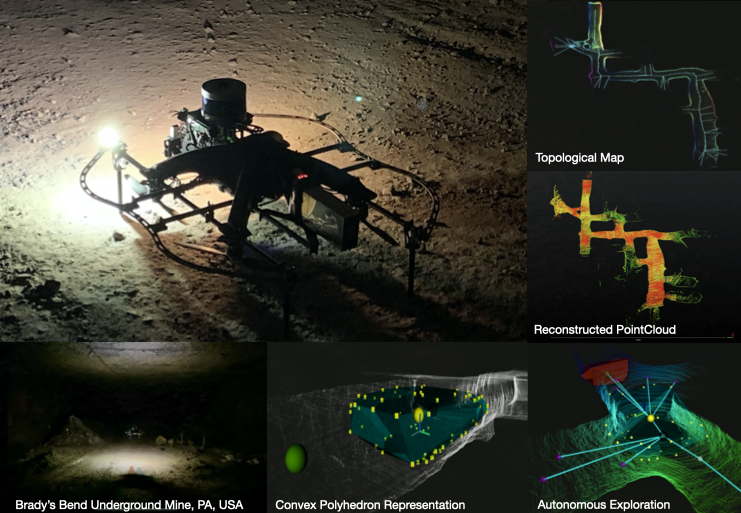}
\caption{An instance of autonomous aerial exploration mission using the proposed graph-based topological exploration planning inside an underground environment located in Armstrong County, Pennsylvania, USA.}
\label{fig:convexhull}
\end{figure}

\noindent Furthermore, a field experiment is conducted to attest to the practical effectiveness of this approach. The main contributions of this paper are summarized as follows:

\begin{enumerate}
    \item A real-time approach that builds a topological representation of 3D environments by utilizing convex polyhedron geometry so as to lower planning redundancy and reduce computational cost.
    
    \item An exploration planning framework on the graph-based topological map which allows ``short-cutting" attempts through uncertain (unknown or partially known) regions, to reduce backtracking maneuvers and improve overall exploration efficiency.
    
    \item Benchmark comparisons of overall exploration performance and efficiency in simulated environments, as well as real-world experiments that validate the applicability and robustness of our proposed method.
    
\end{enumerate}

\section{Related Works}
The frontier-based method was first proposed by \cite{613851}, which utilized a greedy approach that directs the robot to the closest boundary of known and unknown space to extend the region of known area. In recent works, sampling techniques have been adopted  \cite{8981594}, \cite{8968151}, \cite{8202319}, \cite{7487281} due to their property of probabilistic completeness in obtaining exploration goals. In the work of \cite{7487281}, a finite iteration random tree was grown in known free space, and the candidate nodes on the tree were prioritized for execution using volumetric gain along their branches. A recent subterranean work  \cite{8981594} adopted a greedy search strategy on RRT \cite{LaValle1998RapidlyexploringRT} expansion with two planning layers. A local layer encourages consistent exploration within a fixed dimensional local space, while a global layer is responsible for re-directing the robot to a new branch after a local dead-end has been met. Despite the encouraging results, however, these sampling techniques demand high computational costs in 3D space and yield highly redundant world representation that hinders the planning efficiency. Additionally,  due to the randomness and influx of information during exploring,  those ``optimal" selections could easily become obsolete, and then require high-frequency replanning to account for it, which hinders the overall exploration efficiency. 
On the other hand, research efforts have been made to incorporate high-level intent in exploration \cite{doi:10.1002/rob.20130}, \cite{7989297}, \cite{8098718} to reduce planning redundancy and enable more consistent exploration maneuvers. The pioneering work of \cite{doi:10.1002/rob.20130} represented the underground mine as a topological map of intersections, in which a ground platform was deployed to explore edges connecting those intersection nodes. Similarly, a more recent work \cite{7989297} utilized a contour-based method to divide 2D space into topological segmentations representing higher-level tasks, which was then utilized in their following work \cite{8098718} to build a topological map and direct the exploration process. However, those methods are either designed to handle a very specific scenario or hard to be extended and utilized in three-dimensional tasks in real-time. 

In this work, we presents a real-time approach to estimate and separate known 3D space with multiple convex polyhedrons. The enclosed geometry not only enables clean and clear frontier space extraction but also helps group the frontier space into distinctive exploration regions, which typically translate to high-level exploration destinations, creating a sparse topological map in 3D environments that enables efficient exploration. 

\section{Problem Statement}
Consider a robot deployed in an unknown environment with a bounded 3D space $V \subset \mathbb{R}^3$. As the robot moves around, the faithfully covered environments by its onboard sensor(s) are identified as known space (regions) $V_{known}$, while some hollow, narrow, and inaccessible regions are denoted as residual regions $V_{res}$ that are not expected to be covered. Then, the problem is considered fully solved when $V_{known} = V$ \textbackslash  $V_{res}$, while the exploration planning process is defined as determining a sequence of positions $\left\{{\xi_i}\right\} \in \mathbb{R}^3$ to be executed that could extend the $V_{known}$ to achieve $V_{known} = V$ \textbackslash  $V_{res}$, and is subject to vehicle dynamic constraints as well as limited operation time to cover space $V$ given the sensor capacity inside the environment.

\begin{figure}[h!]
\centering
\includegraphics[scale=0.2]{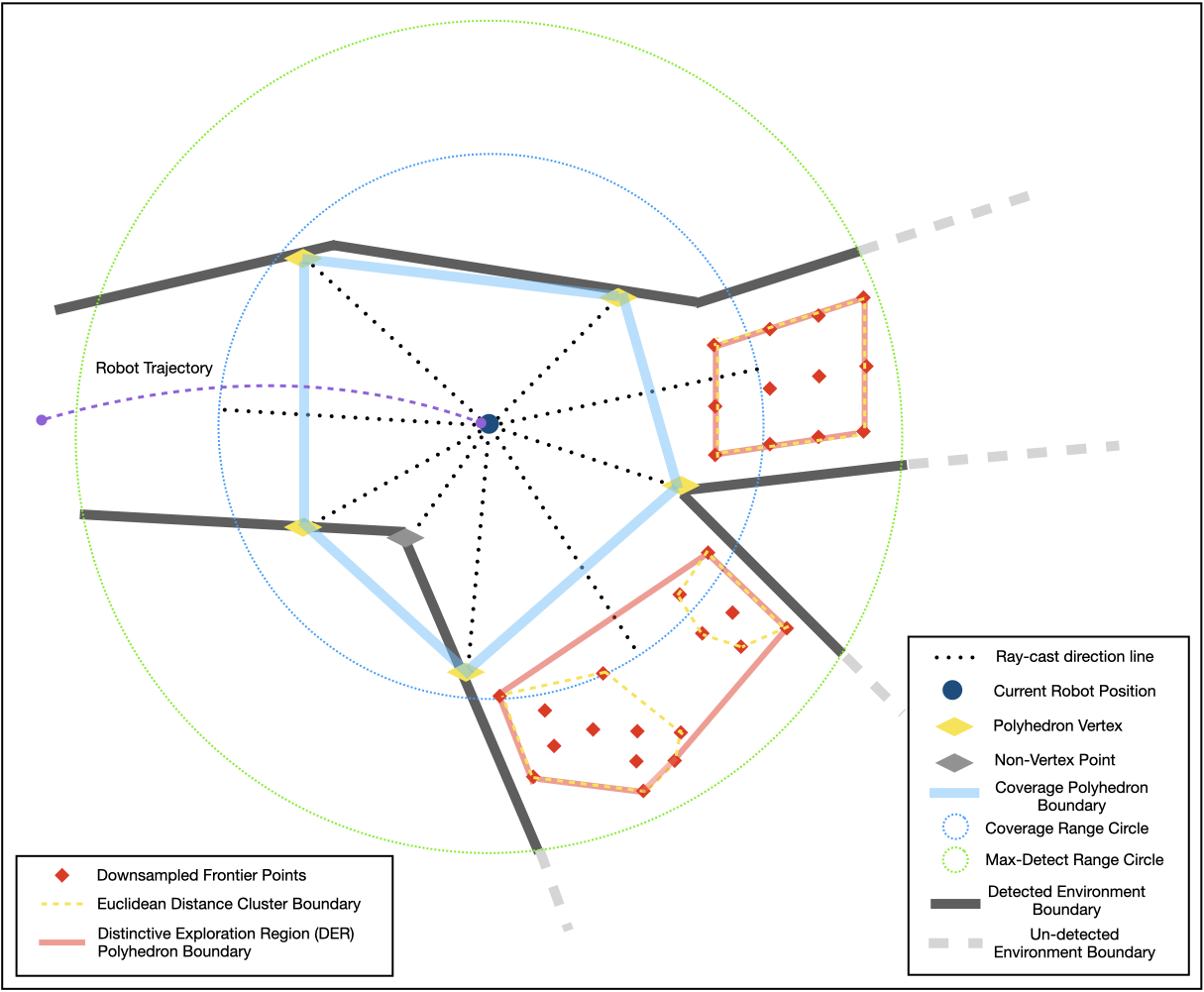}
\caption{Illustration of the proposed polyhedron (shown in 2D as a polygon) space representation. The blue lines indicate the boundary of the convex polyhedron representing the current sensor coverage region, which is generated from the vertices (yellow rhombus) returned by ray-casting (black dot line) in sampled directions. The dot circles illustrate the detection sensor range (green) and coverage range (blue). The distinctive exploration regions (red polygons) are generated from down-sampled frontier points (red dots) captured from raw sensor inputs.}
\label{fig:convexhull}
\end{figure}

\section{Proposed Approach}
\subsection{World Representation}
\subsubsection{Known World Representation}
As defined in \cite{613851}, a frontier is the boundary between known and unknown regions. In three-dimensional space, however, known space observations are often incomplete due to the sensor field-of-view, resolution, and occlusions \cite{6225146}. Thus, a discretized representation (e.g voxel map) of the known world often yields scattered frontiers that lead to poor exploration performance. In this work, a continuous representation of the three-dimensional regions is proposed using convex polyhedrons. To generate such representation, the algorithm is given inputs of the robot position $p$, coverage sensor range $\zeta_{coverage}$ and accumulated sensor inputs. To start, the algorithm first evenly samples n directions $\left\{\tau_{i, i \in 1,2,...,n}\right\}$ in 3D space, then casts rays originating from robot position $p$ on each direction $\tau_i$, which returns reflected surface points $\left\{P_{j,j \in 1,2,...,k}\right\}$ within $\zeta_{coverage}$. A coverage polyhedron $\Lambda_{coverage}$ is then generated from $\left\{P_{j,j \in 1,2,...,k}\right\}$ (illustrated in Figure \ref{fig:convexhull}) using the incremental generation method introduced in \cite{computegeo}. Note that the space contained by a coverage polyhedron $\Lambda_{coverage}$ is used to estimate the coverage region $V_{coverage}$ with the relationship $V_{coverage} \subseteq \Lambda_{coverage}$ , and hereby the union of $\Lambda_{coverage}$ contains the known world $V_{known}$ with the relationship $V_{known} \subseteq \Lambda_{coverage}^{1} \cup \Lambda_{coverage}^{2} \dots \Lambda_{coverage}^{m}$. Note that $V_{known}$ equals to the union of all $\left\{ V^{m}_{coverage}\right\}$.

\subsubsection{The Frontier Regions}
The enclosed geometry of the polyhedral estimation enables robust extraction of frontier space, defined as $V_{frontier} = V_{detect}$ \textbackslash  $V_{known}$, with $V_{frontier}$, $V_{detect}$, $V_{known} \subset \mathbb{R}^{3}$. Specifically, the detected region $V_{detect}$ corresponds to all previously observed space within detection sensor range $\zeta_{detect}$ while the coverage range $\zeta_{coverage}$ represents the range within which the environment can be faithfully covered. Note that $V_{detect} \supseteq V_{known}$, and
the tunable gap between $\zeta_{detect}$ and $\zeta_{coverage}$ affects the quantity and quality of frontier space extraction and requires engineering tuning under different environments with different sensors. In the theoretical case where  $\zeta_{detect} = \zeta_{coverage}$, $V_{frontier}$ would degenerate to the surrounding surface of the known space with $V_{known} = V_{detect}$.

\subsubsection{Distinctive Exploration Regions (DER-s)}
Frontier space $V_{frontier}$ is divided into distinctive regions $V_{DER}$-s which are estimated by corresponding convex polyhedrons $\Lambda_{DER}$-s, similar to coverage regions with the relationship $V_{DER} \subseteq \Lambda_{DER}$ and hereby with $V_{frontier} \supseteq \Lambda_{DER}^{1} \cup  \Lambda_{DER}^{2} \dots \Lambda_{DER}^{h}$, given that $V_{frontier}$ equals to the union of all $\left\{ V^{h}_{DER}\right\}$. By achieving distinctiveness, each $V_{DER}$ could typically represent an entire enclosed space such as a room or corridor. Thus, using $V_{DER}$-s as exploration candidates would generally encode each exploration step with high-level intent such as moving from a corridor to another room. This alleviates the demand for high-frequency re-planning and yields a more efficient and consistent behavior. 

\begin{figure}[h!]
\centering
\includegraphics[scale=0.18]{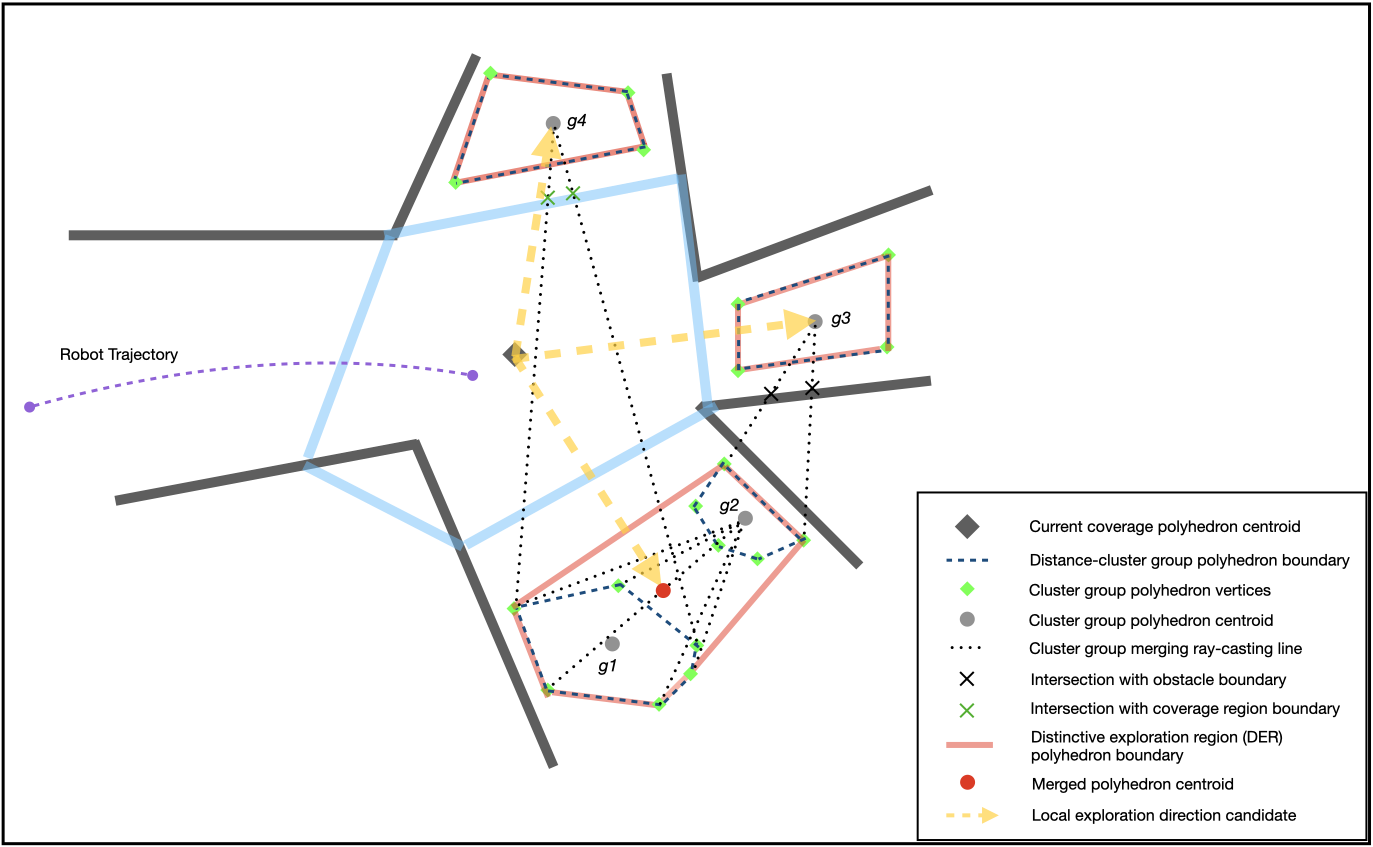}
\caption{Illustration of DERs generation and mutual visibility merging strategy. The frontier regions (represented as points) are generated by subtracting coverage regions from detected regions. These regions are then clustered into different euclidean-distance-cluster groups $g1$, $g2$, $g3$, $g4$, represented by individual polyhedrons (black dash polygons). Note that $g1$ and $g2$ are merged into one distinctive region (bottom red polygon) since their vertices can be connected from each other's geometry centroid. Meanwhile, however, $g2$ and $g3$ are considered distinctive since their connections are blocked by obstacles in between. Similarly, $g1$ and $g4$ are considered distinctive since their connections are interrupted by the coverage polyhedron (blue polygon).}
\label{ fig:distinctive_exploration_region}
\end{figure}

In this work, the distinctiveness of $V_{DER}$-s is achieved by examining the mutual visibility of their polyhedron estimations $\Lambda_{DER}$-s. Specifically, any two $V_{DER}$-s whose polyhedral vertices and centroid maintain mutual line-of-sight, without being occluded by obstacles or interrupted by coverage regions, will be merged as one DER (see Figure \ref{ fig:distinctive_exploration_region} for illustration). The method to generate and update $V_{DER}$-s is presented in Algorithm \ref{alg:algorithm1} which starts by employing euclidean-distance-based clustering method on a down-sampled points representation $F$ of frontier space, separating $F$ into a group of spatially dis-contiguous clusters $\left\{g_i\right\} \in G$. Then for each cluster $g_i$, a convex polyhedron $\Lambda^{new}$ is generated from all points in cluster $g_i$ to estimate the geometry of a new candidate distinctive region $v^{new}$. Finally, the candidate region $\left\{ v^{new}, \Lambda^{new} \right\}$ will be checked against nearby existing DERs $\left\{v_j^{\prime}, \Lambda_j^{\prime} \right\} \in V^{\prime}$ for mutual visibility. 

\begin{algorithm}
\caption{DERs Generate and Update}
\label{alg:algorithm1}
\begin{algorithmic}[1]
\INPUT Frontier Points $F$, Existing Distinctive Regions $V_E$
\OUTPUT Updated Distinctive Regions $V_{Update}$
\State $G \leftarrow$ EuclideanDistanceCluster($F$)
\State $g \leftarrow$ $G$.begin()
\ForAll{$g \in G$} 
  \State $\left\{ v^{new}, \Lambda^{new} \right\}$ $\leftarrow$ GenerateDistinctiveRegion($g$)
  \State $V^{\prime} \leftarrow$ NearbyExistDistinctiveRegions($V_E$, $v^{new}$)
  \State $\left\{v^{\prime}, \Lambda^{\prime} \right\}$ $\leftarrow$ $V^{\prime}$.begin()
    \State merge\_flag $\leftarrow$ \textbf{false}
    \While{$\left\{ v^{\prime},  \Lambda^{\prime} \right\} \neq V^{\prime}$.end()}  
        \If{HasMutualVisibility($\Lambda^{new}$, $\Lambda^{\prime}$)}
            \State $\left\{v^{\prime}, \Lambda^{\prime} \right\} \leftarrow$ MergeRegions($\Lambda^{new}$, $\Lambda^{\prime}$)
            \State merge\_flag $\leftarrow$ \textbf{true}
            \Break
        \EndIf
    \State$\left\{v^{\prime}, \Lambda^{\prime} \right\} \leftarrow \left\{v^{\prime}, \Lambda^{\prime} \right\}.next()$
    \EndWhile
\If{\textbf{not} merge\_flag}
    \State $V_E.add(\left\{v^{new}, \Lambda^{new} \right\})$
\EndIf
\EndFor
\State $V_{Update} \leftarrow V_{E}$
\end{algorithmic}
\end{algorithm}

\subsection{Graph-based Topological Map}
\begin{figure}[h!]
\centering
\includegraphics[scale=0.22]{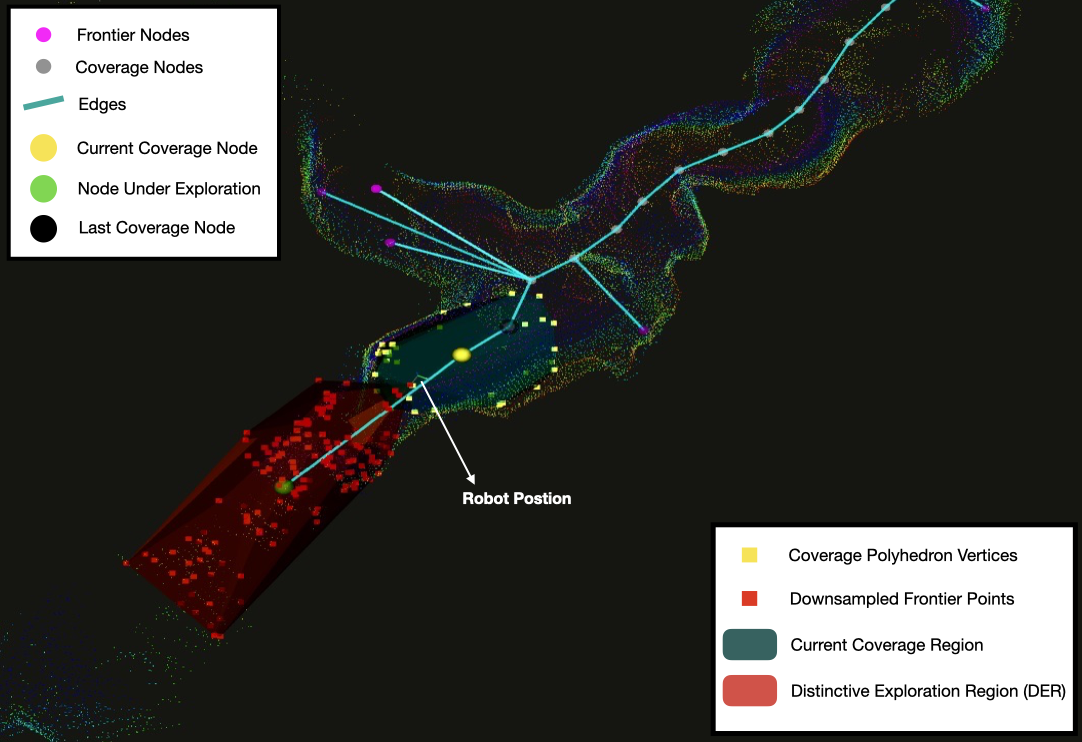}
\caption{A visualized instance of an evolving graph-based topological map during exploration process in the simulated cave environment provided for DARPA SubT Challenge Virtual Competition, Cave Circuit \cite{subt}. The polyhedron (red) represents the DER currently under exploration. The polyhedron (emerald) represents the current coverage region.}
\label{fig:topological map}
\end{figure}

A graph-based topological map, shown in Figure \ref{fig:topological map}, is built with coverage regions and DERs using Algorithm \ref{alg:algorithm2}, which represents coverage regions $\left\{ V^{m}_{coverage}\right\}$ and DER regions $\left \{V^h_{DER} \right \}$ as graph nodes $\left\{N^m_{c}\right\}$ and $\left\{N^h_{f}\right\}$ respectively, with their connectivity represented as edges. Specifically, as the robot moves out of the current coverage region, a new coverage node $N^{p}_{c}$ will be created from the current robot position $p$ to represent the new covered environment and estimated as a convex polyhedron.  To reduce redundancy and maintain distinctiveness among graph nodes, however, if the new node $N^{p}_{c}$  has its polyhedron's geometric centroid contained by another polyhedron of an existing nearby coverage node $N_c^{\prime}$, it will be merged into the node $N_c^{\prime}$. For frontier nodes, similarly, if an existing frontier node $N_{f}^{\prime}$ with its centroid contained inside the space represented by $N^{p}_{c}$, it will then be deleted from the topological map. 

Meanwhile, edges in a topological map can be categorized into two types based on the nodes $N \in \{N_{c}, N_{f}\}$ that they're bridging:  $N_{c}$ - $N_{c}$ edges represent the adjacent traversability between two coverage regions which are established by examining the spatial overlap between contiguous $V_{coverage}$. On the other hand, $N_{c}$ - $N_{f}$ edges translate to  potentially-traversable-paths $\eta$. And for each $N_f$, it only connects to the coverage node $N_{c}^*$ with the lowest-cost-path $\eta^*$ that links to it. Specifically, each $\eta$ is generated using a \emph{Cylinder-Astar-Path-Planning} algorithm which is a modified $A^{*}$ planning algorithm \cite{4082128} with cylindrical space constraint, beyond whose boundary a state will not be expanded. Such cylinder approximates the ability of trajectory planner to workaround local obstacles, and has axis in line with $N_{c}$-$N_{f}$ and tunable radius r proportional to \emph{Distance}($N_{c}$, $N_{f}$). 
\begin{algorithm}
\caption{Topological Exploration Map}
\label{alg:algorithm2}
\begin{algorithmic}[1]
\INPUT Robot Position $p$, Distinctive Regions $V_{f}$, Graph $G$
\OUTPUT: Updated Graph $G$
\State merge\_flag $\leftarrow$ \textbf{false}
\State $N^{p}_{c} \leftarrow$ CoverageNodeFromPosition($p$) 
\State $G$.DeleteOverlapFrontierNodes($N^{p}_{c}$)
\State $\Gamma_{c}\leftarrow$ SurroundingCoverageNodes($N^{p}_{c}$, $G$)
\ForAll{$N_{c}^{\prime} \in\Gamma_{c}$}
    \If{CentroidOverlap($N^{p}_{c}$, $N_{c}^{\prime}$)}
        \State merge\_flag $\leftarrow$ \textbf{true}
        \State $N^{\prime} \leftarrow$ MergeNode($N^{p}_{c}$, $N_{c}^{\prime}$)
        \Break
    \EndIf
\EndFor
\If{\textbf{not} merge\_flag }
    \State $G$.addNode($N^{p}_{c}$)
    \State $\Gamma_{overlap} \leftarrow$  SpatialOverlapNodes($N^{p}_{c}$, $\Gamma_{N}$) 
    \State $G$.addEdgeToNodes($N^{p}_{c}$, $\Gamma_{overlap}$)
\EndIf
\ForAll{$v \in V_{f}$}
    \State$N_{f} \leftarrow$ FrontierNodeFromRegion($v$)
    \State $N_{c}^{*} \leftarrow \displaystyle \argmin_{N_{c}^{\prime} \ \in \ \Gamma_{c}}$ Cost(CylinderAstarPath($N_{c}^{\prime}$, $N_{f}$))
    \State $\eta^{*} \leftarrow $  CylinderAstarPath($N_{c}^{*}$, $N_{f}$)\hfill 
    \If{Cost($\eta^{*}$) $< \infty$}
        \State $G$.addNode($N_{f}$)
        \State $G$.addEdge($N_{f}$, $N^{*}$)
    \EndIf
\EndFor

\end{algorithmic}
\end{algorithm} 
 
\/*
\begin{algorithm}
\caption{Algorithm3}
\label{alg:algorithm3}
\begin{algorithmic}[1]
\INPUT Start Node $N_{start}$, Goal Node $N_{goal}$
\OUTPUT Path $\eta$ connecting $N_{start}$ and $N_{goal}$
\State $S_{start} \leftarrow$ StateFromNode($N_{start}$)
\State $S_{goal} \ \leftarrow $ StateFromNode ($N_{goal}$)
\State $\Gamma_{open} \leftarrow \{ S_{start}\}$ \hfill  //sorted by estimate cost-to-goal
\State $\Gamma_{close} \leftarrow \emptyset$
\While{not $\Gamma_{open}$.empty()} 
    \State $S_{curr} \leftarrow $ LowestCostState($\Gamma_{open}$) 
    \If{Dist($S_{curr}$, $S_{goal}$)  $\leq \epsilon$}
        \State \Return BackTracking($S_{curr}$, $S_{start}$)
    \EndIf
    \ForAll{$S_{next} \in$ Neighbors($S_{curr}$)}
        \If {$S_{next} \in \Gamma_{close}$} \CONTINUE
        \EndIf
        \If {not CollisionFree($S_{next}$)} \CONTINUE
        \EndIf
         \If {not InCylinder($S_{next}$)} \CONTINUE \hfill  
        \EndIf
        \If{not $S_{next}$.isVisited()}
         $S_{next}$.cost $\leftarrow \infty$
        \EndIf
        \State $\Gamma_{close}$.insert($S_{curr}$)
        \State $\Gamma_{open}$.erase($S_{curr}$)
        \State $C_{ref} \leftarrow S_{curr}$.cost + Dist($S_{curr}$, $S_{next}$)

        \If{$C_{ref} < S_{next}$.cost}
        \State UpdateNodeState($S_{next}$, $C_{ref}$)
        \State $\Gamma_{open}$.insert($S_{curr}$)
        \EndIf
    \EndFor
\EndWhile 
\State \Return EmptyPath()
\end{algorithmic}
\end{algorithm}*/

\subsection{Graph Based Exploration Planning Strategy}
\subsubsection{Local Topology Exploration Strategy}
This work has adopted a similar bifurcated planning strategy as proposed in \cite{8981594}: combining a local and a global exploration strategy. Locally, on the topological map, the frontier nodes $\left\{\hat{N}^{k}_{f}\right\}$ that connect to the current coverage node are evaluated and assigned with a normalized exploration score based on Equation \eqref{eq:LocalExploreScore}. The robot is then directed greedily to the frontier node with the highest local exploration score.

Equation \eqref{eq:LocalExploreScore} is composed of three normalized factors of range $(0, 1]$, with the normalization operations denoted by $``*"$. Specifically, the first factor $\emph{Volume}^*$ is computed with the spacial volume of $\hat{N}_{f}$ to award largely unexplored distinctive region; the second factor $\emph{Distance}^*$ is computed with Euclidean distance between centroid of $\hat{N}_{f}$ and current robot position $p$, to encourage longer travel distance and its consequent higher potential information gain along the path; the final factor $\emph{Direction}^*$ is calculated by projecting the new exploration direction $\hat{\psi}$ to the current exploration direction $\psi$, to encourage consistent heading and prevent sudden back-and-forth maneuvers.

\begin{equation}\label{eq:LocalExploreScore}
    \begin{split}
& \textbf{LocalExploreScore}(\hat{N}_f) \ = \\ &\emph{Volume}^*(\hat{N}_f) \cdot  \emph{Distance}^*(\hat{N}_f, p) \cdot \emph{Direction}^*(\hat{\psi}, \psi)   
    \end{split}
\end{equation}

\subsubsection{Global Exploration Strategy with Adaptive Planning through Uncertainty}
On the other hand, a global graph planner will kick in once a local ``dead-end" is met i.e. a position with no directly connected frontier node, to re-direct the robot to a global frontier node $N_{f}$ with the highest global exploration score assigned by Equation \eqref{eq:GlobalExploreScore}. Note that, due to incomplete exploration, the topological map may lack edges between regions that could have physical connections through unexplored or partially explored areas. Thus, the planner adaptively searches and updates the path to the destination frontier with incoming information while exploring. That allows the attempts of the robot to travel through unexplored areas without backtracking large visited areas during global re-direction. The planning-through-uncertainty strategy described in Algorithm \ref{alg:uncertainty_planning} is built on $A^{*}$ planning algorithm \cite{4082128}. Specifically, for each $N^*$ expanded, all adjacent nodes inside a tunable distance $\sigma$, with or without an existing edge, will be considered as potential-next-states. The adjacent nodes without a connecting edge to node $N^*$ will have their cost-to-connect estimated by an uncertainty penalizing factor $\gamma$, times the path cost returned by the aforementioned \emph{Cylinder-Astar-Path-Planning} method.

\begin{equation}\label{eq:GlobalExploreScore}
    \begin{split}
 &\textbf{GlobalExploreScore}(N_f) \ = \\ &\emph{Volume}(N_f) \cdot  \frac{1.0}{\emph{Distance}(N_f, p) + 1.0}   
    \end{split}
\end{equation}

\begin{algorithm}
\caption{Global Planning through Uncertainty}
\label{alg:uncertainty_planning}
\begin{algorithmic}[1]
\INPUT Graph $G$, Goal Node $N^{*}_{f}$,  Current Position Node $N_{c}^{p}$
\OUTPUT Path $\eta(N_{c}^{p}, N^{*}_{f})$ 
\State $\Gamma_{open} \leftarrow \{ N_{c}^{p}\}$ 
\State $\Gamma_{close} \leftarrow \emptyset$
\While{not $\Gamma_{open}$.empty()}
    \State $N^{*} \leftarrow $ LowestCostNode($\Gamma_{open}$) 
    \If{$N^{*}$ = $N^{*}_{f}$}
        \State \Return PathTrace($N^{*}_{f}$, $N_{c}^{p}$, $G$)
    \EndIf
    \State $\Gamma_{close}$.insert($N^{*}$)
    \State $\Gamma_{open}$.erase($N^{*}$)
    \State $\Gamma_{adj} \leftarrow$  AdjacentNodesInDistance($G$, $N^{*}$, $\sigma$)
    \ForAll{$N_{adj} \in \{\Gamma_{adj} \setminus \Gamma_{close}\}$}
        \If{HasEdge($N_{adj}$, $N^{*}$)}
            \State $C_{ref} \leftarrow N_{adj}$.cost() + Distance($N^{*}$, $N_{adj}$)
        \Else
            \State $\eta \leftarrow$ CylinderAstarPath($N^{*}$, $N_{adj}$)
            \State $C_{ref} \leftarrow N_{adj}$.cost() + $\gamma$ * Cost($\eta$)
        \EndIf
        \If{$C_{ref} <  N_{adj}$.cost()}
            \State UpdateNodeCostAndParent($N_{adj}$, $C_{ref}$, $N^{*}$)
            \State $\Gamma_{open}$.addNode($N_{adj}$)
        \EndIf
    \EndFor
\EndWhile

\end{algorithmic}
\end{algorithm}

\section{Experiments and Results}
\subsection{Evaluation Metrics}
To benchmark and compare between different exploration planning methods, the overall exploration performance is evaluated by the explored or detected volume$(m^3)$ with a certain sensor model over a certain time$(s)$, which directly reflects the exploration outcomes. Additionally, the exploration efficiency is defined as the median of exploration rates ($m^3/s$) during the whole exploration process to evaluate the quality of exploration steps and overall planning maneuvers.

\subsection{Simulation Based Evaluation}

\begin{figure}[h!]
    \begin{subfigure}{0.5\textwidth}
        \centering
        \includegraphics[scale=0.13]{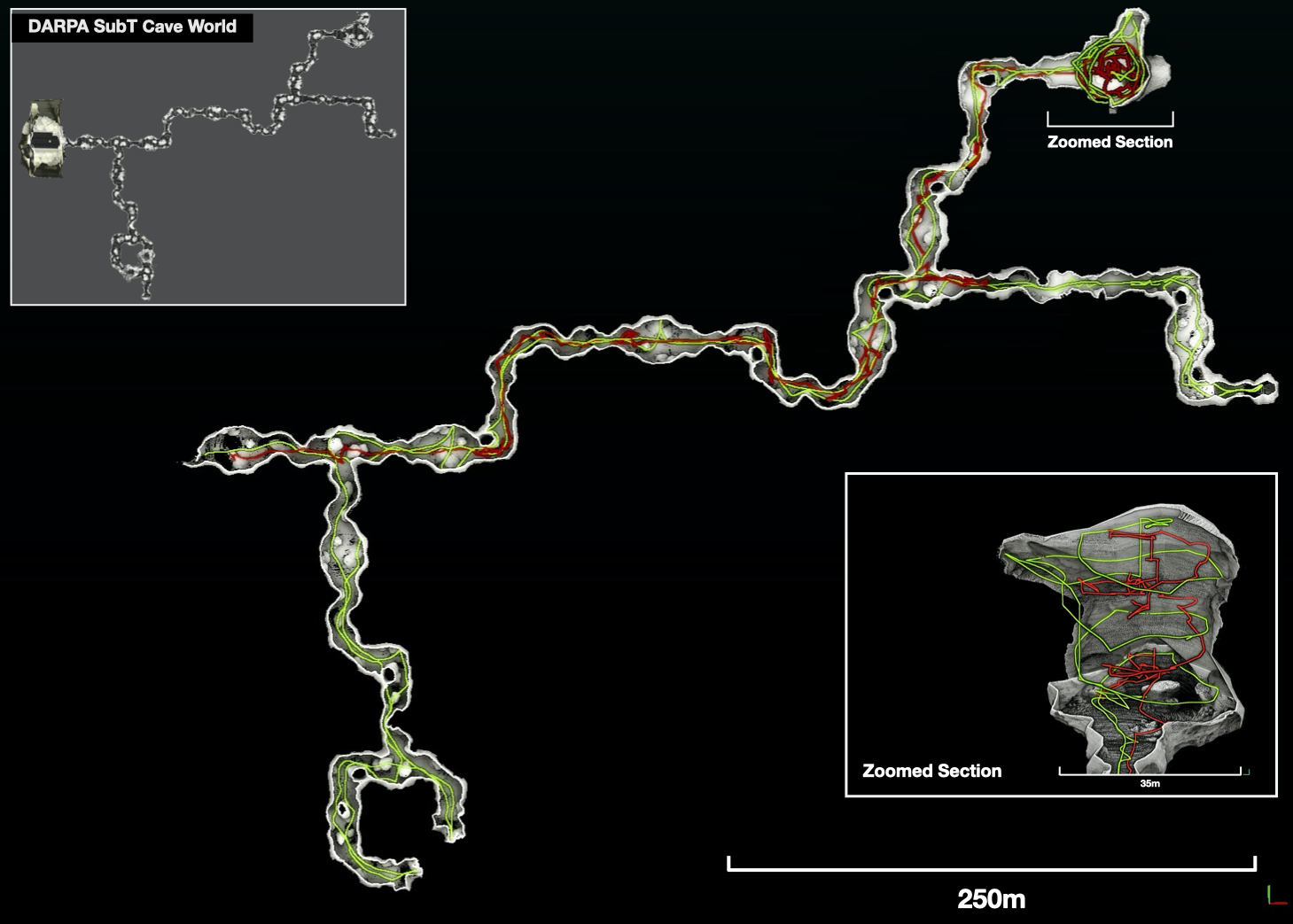}
        \caption{}
    \end{subfigure}
    
    \begin{subfigure}{0.5\textwidth}
        \centering
        \includegraphics[scale=0.45]{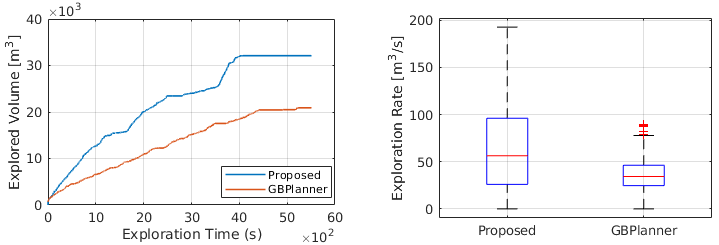}
        \caption{} 
    \end{subfigure}

    \caption{An instance of DARPA \cite{subt}  cave world test result. (a) Visualization of the reconstructed point-cloud map explored by the proposed method (green trajectory). The other exploration trajectory (red) is taken by the baseline method \cite{8968151}, considered as state-of-the-art. (b) The comparative result for explored volume over time and statistics of exploration rates during the process. } 
    \label{fig:simulated_sutb_cave}
\end{figure}

\begin{figure}[h!]
    \begin{subfigure}{0.5\textwidth}
        \centering
        \includegraphics[scale=0.121]{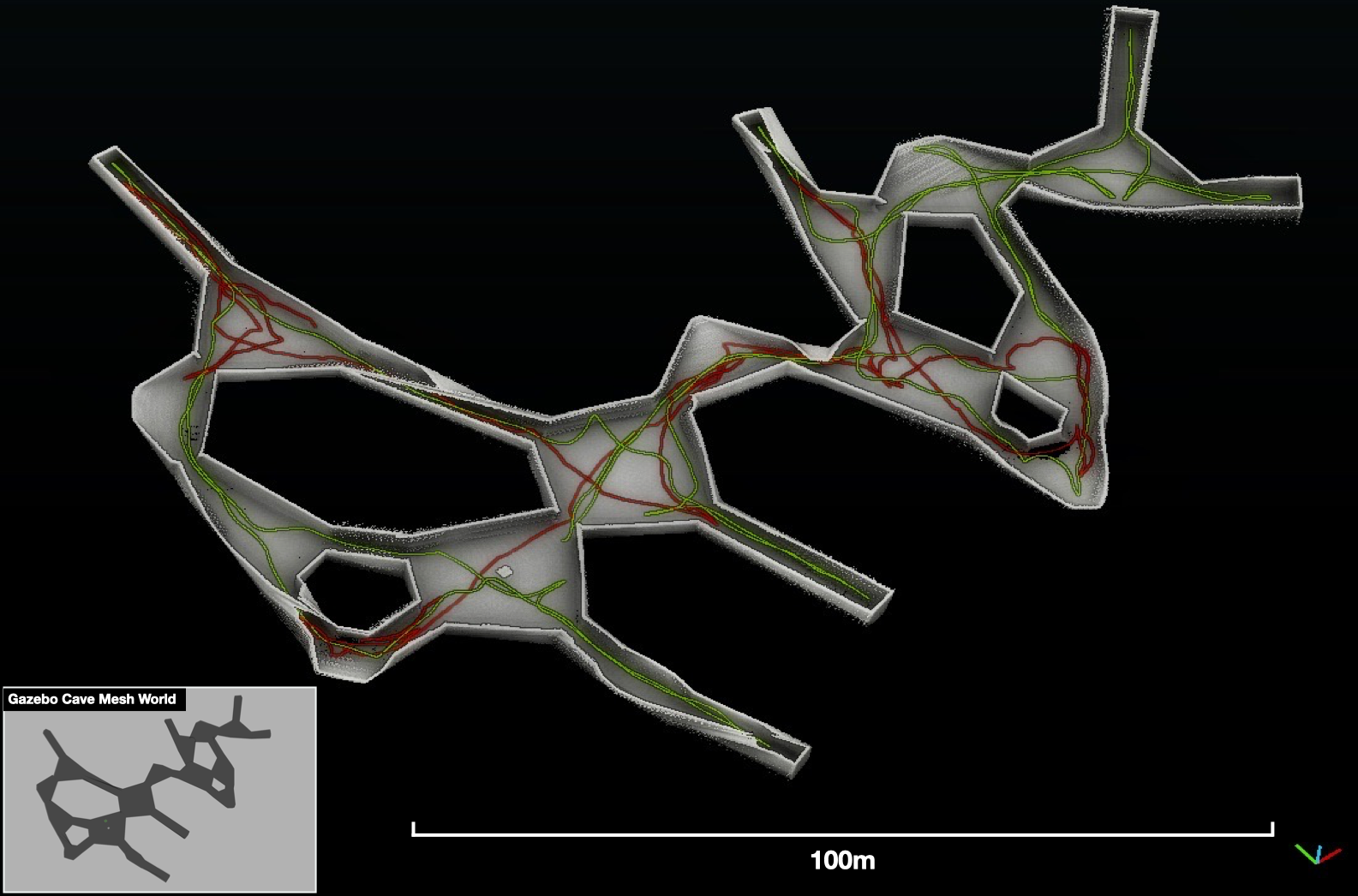}
        \caption{}
    \end{subfigure}
    
    \begin{subfigure}{0.5\textwidth}
        \centering
        \includegraphics[scale=0.39]{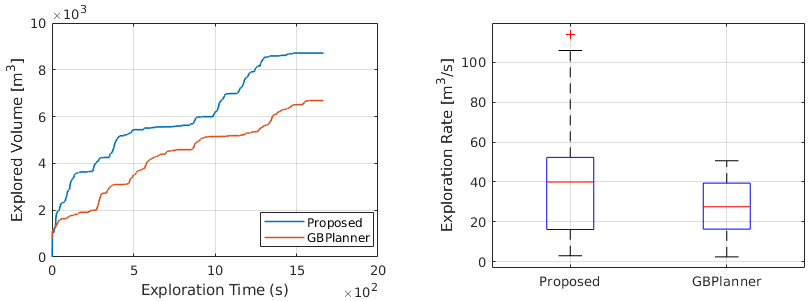}
        \caption{} 
    \end{subfigure}
    \caption{An instance of Gazebo cave mesh world test result. The figure shares the same layout as Fig. \ref{fig:simulated_sutb_cave}.} 
    \label{fig:simulated_random_cave}
    
\end{figure}

In this section, the performance of our proposed method is compared to the baseline method \cite{8968151}, considered as state-of-the-art, in 3D simulated environments: (1) world model provided by DARPA subterranean virtual competition \cite{subt} (2) cave mesh world manually generated with ROS-Gazebo. The results are shown in Figure \ref{fig:simulated_sutb_cave} and Figure \ref{fig:simulated_random_cave} respectively. All the experiments were conducted within a bounded time unless the planners reported a completion status. A LiDAR model is deployed with the following specification: The detection sensor range $\zeta_{detect}$ is set to 15 meters and the coverage range $\zeta_{coverage}$ is set to 10 meters, with horizontal and vertical field of view being $F_H = 360^{\circ}$ and $F_V = 30^{\circ}$ respectively. The maximum flight speed of the robot is set to $0.75 m/s$. In both scenarios, our method achieves higher exploration efficiency and explored-volume than state-of-the-art. Besides, our proposed method shows more consistent exploration maneuvers with fewer back-and-forth behaviors, as indicated by the trajectories.

\begin{figure}[h!]
\centering
\includegraphics[scale=0.18]{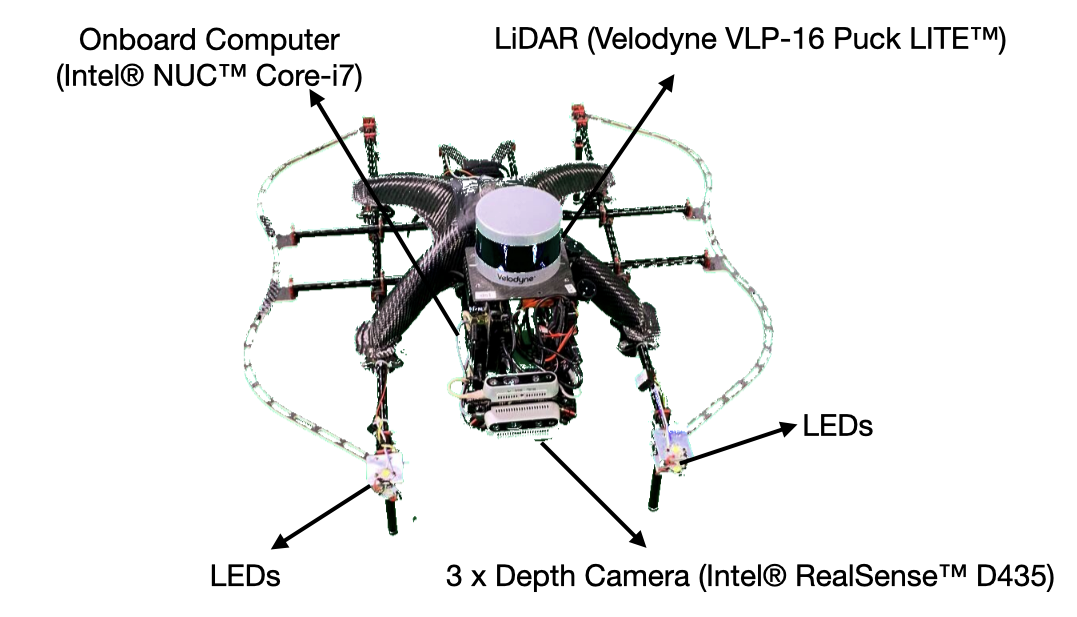}
\caption{The aerial robot configuration deployed in the real-world experiment.}
\label{fig:drone}
\end{figure}

\subsection{Experimental Evaluation}

To further investigate the applicability of the proposed method to empower efficient and robust exploration in complex real-world environments, a field experiment was conducted at ``Brady's Bend", an inactive limestone mine in Armstrong County, Pennsylvania, USA. This underground mine consists of wide, long halls with smaller corridors branching off from the main structures and connecting with each other. The whole exploration mission is fully autonomous. Below, details of the deployed aerial robot and the field evaluation result are provided.

\subsubsection{System Overview}
A quadrotor aerial robot, as shown in Figure \ref{fig:drone}, is utilized in this real-world underground exploration experiment. The aerial robot is integrated with a Velodyne Puck LITE LiDAR, providing horizontal and vertical fields-of-view of $F_H = 360^{\circ}$ and $F_V = 30^{\circ}$, with maximum range of $100m$. The LiDAR sensor outcomes provide environment information and are utilized for state estimation by a lidar-based odometry and mapping algorithm \cite{LOAM}. Additionally, three Intel Realsense depth cameras (up, down, and forward) are equipped along with LED lights to provide more detailed information of surrounding obstacles to enhance flight safety. A proportional–integral–derivative (PID) controller is employed and responsible for guiding the robot along local trajectories generated by a motion-primitives planner that selects feasible paths based on current robot states and surrounding obstacles from a pre-built trajectory library. All on-board tasks are executed in real-time using an Intel NUC Core-i7 computer mounted inside the robot. The maximum speed is set to $1.0m/s$ during the flight.

\subsubsection{Autonomous Exploration of Subterranean Environment}
\begin{figure}[h!]
    \begin{subfigure}{0.5\textwidth}
        \centering
        \includegraphics[scale=0.14]{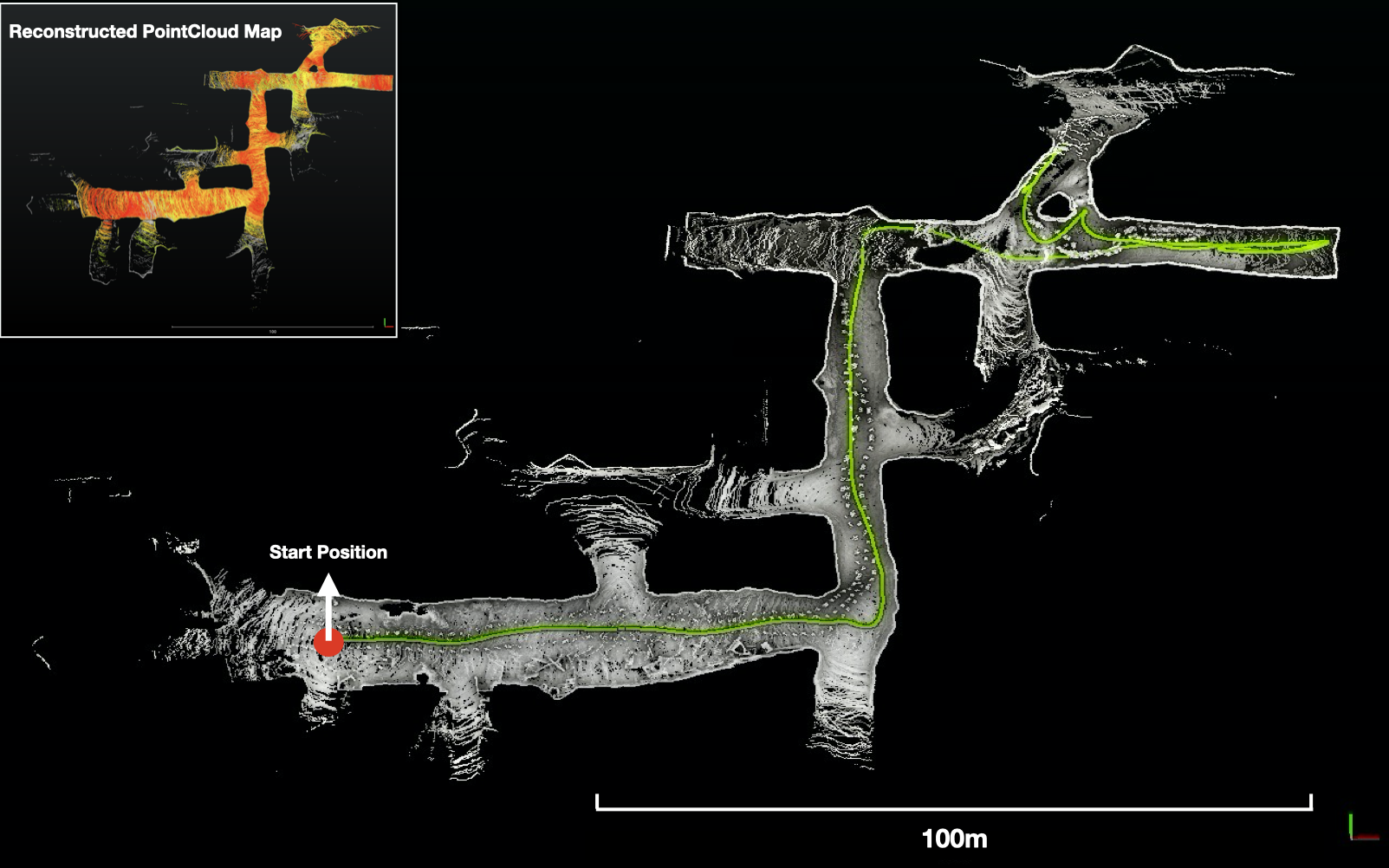}
        \caption{} 
    \end{subfigure}
    \begin{subfigure}{0.5\textwidth}
        \centering
        \includegraphics[scale=0.4]{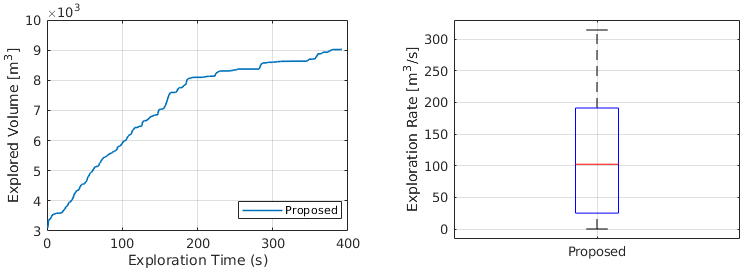}
        \caption{} 
    \end{subfigure}
    \caption{Real-world flight test in Brady's Bend underground environment. (a) Visualization of the reconstructed point-cloud map as well as the trajectory (green) (b) The overall exploration performance and efficiency results.}
    \label{fig:experiment_performance.png}
\end{figure}

In this field experiment, the detection sensor range $\zeta_{detect}$ of the exploration planner is limited to $15m$ as the detection beyond that range becomes too sparse to be used. The coverage sensor range $\zeta_{coverage}$ is set to $10m$ to ensure the faithful and complete information coverage inside that range, and also to leave enough gap between detection range to extract frontier space $V_{frontier}$. The aerial robot was deployed at one side of the main hall structure and explored autonomously until the end of its battery allowance. Indicated by the exploration trajectory (green), shown in Figure \ref{fig:experiment_performance.png}(a), the experiment demonstrates the planner's ability to maintain a high exploration rate by exploring along with the main directions of the halls and corridors with consistent maneuvers, and redirect the robot to explore a new branch after a local ``dead-end" has been met. Additionally, the exploration mission lasted for around $390s$ and the robot flied around $310m$, which results in an overall average exploration speed of $0.8m/s$, versus the maximum speed limitation, $1.0m/s$. That again indicates the planner's ability to make consistent decisions and enable high exploration rate, as results shown in Figure \ref{fig:experiment_performance.png}(b).

\section{Conclusion and Future Work}
This paper proposes a planning approach for large-scale exploration. The 3D space is estimated and separated by convex polyhedrons whose geometric information is then utilized to extract distinctive regions and form a sparse graph-based topological map. Furthermore, a bifurcated planning strategy is adopted to direct the robot to explore towards local distinctive regions, while globally ensuring continuous exploration and short-cutting through uncertain areas. The proposed method is compared in simulated environments and shows better overall exploration performance and higher exploration efficiency than the state-of-the-art counterpart. Besides, a real-world experiment is conducted to verify the effectiveness and applicability of this work. Currently, the method uses geometry-based metrics to extract distinctive regions and form a topological map without considering the semantic meanings of the environment. In the future, we plan to have semantic information side with geometry information to better extract distinctive regions. Additionally, we also expect to achieve multi-robot exploration based on topological map sharing to explore the environment collectively and enable much higher exploration efficiency.   

\section*{Acknowledgment}

Approved for public release; distribution is unlimited. This research was sponsored by DARPA $\text{HR}00111820044$. Content is not endorsed by and does not necessarily reflect the position or policy of the government.

\bibliographystyle{IEEEtran}
\bibliography{references}

\addtolength{\textheight}{-12cm}   



\end{document}